# ARMOR: Adaptive Resilience Against Model Poisoning Attacks in Continual Federated Learning for Mobile Indoor Localization

## Danish Gufran, Akhil Singampalli, Sudeep Pasricha

Colorado State University, Fort Collins, CO 80523, USA

Abstract: Indoor localization has become increasingly essential for applications ranging from asset tracking to delivering personalized services. Federated learning (FL) offers a privacy-preserving approach by training a centralized global model (GM) using distributed data from mobile devices without sharing raw data. However, real-world deployments require a continual federated learning (CFL) setting, where the GM receives continual updates under device heterogeneity and evolving indoor environments. In such dynamic conditions, erroneous or biased updates can cause the GM to deviate from its expected learning trajectory, gradually degrading internal GM representations and GM localization performance. This vulnerability is further exacerbated by adversarial model poisoning attacks. To address this challenge, we propose ARMOR, a novel CFL-based framework that monitors and safeguards the GM during continual updates. ARMOR introduces a novel state-space model (SSM) that learns the historical evolution of GM weight tensors and predicts the expected next state of weight tensors of the GM. By comparing incoming local updates with this SSM projection, ARMOR detects deviations and selectively mitigates corrupted updates before local updates are aggregated with the GM. This mechanism enables robust adaptation to temporal environmental dynamics and mitigate the effects of model poisoning attacks while preventing GM corruption. Experimental evaluations in real-world conditions indicate that ARMOR achieves notable improvements, with up to 8.0× reduction in mean error and 4.97× reduction in worst-case error compared to state-of-the-art indoor localization frameworks, demonstrating strong resilience against model corruption tested using real-world data and mobile devices.

**Keywords:** Wi-Fi fingerprint, federated learning, continual learning, model poisoning attacks, model corruption, device heterogeneity, temporal dynamics, indoor localization.

## 1. Introduction

The domain of indoor localization is rapidly expanding, driven by advancements in healthcare, asset tracking, indoor navigation systems, virtual reality, smart home automation, and location-based advertising. The indoor positioning and indoor navigation market was valued at USD 11.9 billion in 2024 and is expected to grow significantly, reaching an estimate of USD 29.8 billion by 2028 [1]. Numerous companies are investing heavily in this technology, including IndoorAtlas, Navigine, and Quuppa, which are developing advanced indoor positioning solutions. In addition, Google has integrated indoor maps into Google Maps [2], and Apple is incorporating indoor positioning features within its ecosystem [3]. The ability to accurately determine the position of mobile devices has the potential to revolutionize industries and enhance user experiences [4], [5].

One effective method for indoor localization is based on Wi-Fi received signal strength (RSS) fingerprinting-based systems. Wi-Fi RSS refers to the strength of the Wi-Fi signal received from a Wi-Fi access point (AP) by a mobile device [6]. To create a fingerprint, the RSS values are collected across all available Wi-Fi APs at the same location, and this collective set of values constitutes what is known as a "fingerprint". Such fingerprinting-based solutions have been widely accepted due to the widespread availability of Wi-Fi in indoor locales, compatibility with today's mobile devices, and ability to provide fine-grained location insights [1], [6], [7].

Wi-Fi RSS fingerprinting-based systems typically involve two phases: the offline phase and the online phase [7], [8]. In the offline phase, fingerprints are collected across several reference points (RPs, i.e., locations) within the indoor space. These collected fingerprints are stored in a database, which serves as the initial training data for a machine learning (ML) model. The ML model processes this data by representing each RSS value as a feature [9]. During training, the ML model adjusts its parameters to minimize the prediction error. The model weights are optimized iteratively to establish a robust relationship between the input features (RSS fingerprints) and the output labels (RPs, i.e., locations), enabling the ML model to make accurate predictions during the online phase when deployed on a mobile device. In the online phase, the trained ML model is used to estimate the location of the mobile device based on new fingerprints collected at an unknown RP in the indoor space [9], [10].

There are several factors that can negatively impact the accuracy of the Wi-Fi fingerprinting approach. Fluctuations in RSS during the online phase can violate the feature patterns learned by the ML weights from the offline phase, leading to mispredictions by the ML model. These fluctuations can be caused by the heterogeneity of different mobile devices (differences

Author's email addresses: D. Gufran (danish.gufran@colostate.edu), A. Singampalli (akhil.singampalli@colostate.edu), S. Pasricha (sudeep@colostate.edu).

in hardware, software, and firmware) [11], as well as the temporal environment dynamics of the indoor space [12]—such as reflections of Wi-Fi signals from obstacles (e.g., furniture), multipath fading, and shadowing effect. These fluctuations create a mismatch between the trained feature patterns from the offline phase and the real-time data in the online phase, necessitating frequent updates to the database to maintain accuracy.

To address these challenges, adopting a distributed ML approach can be highly beneficial [13]. This method not only helps in learning the fluctuations caused by device heterogeneity but also maintains user data privacy and reduces the cumbersome data collection process. Federated learning (FL) is a promising solution that can utilize distributed ML to overcome many of the challenges facing traditional localization systems [13], [14]. FL typically consists of a centralized server that hosts a global model (GM). Initially, the GM is constructed during the offline phase. In the online phase, several participating mobile devices, known as clients, receive a copy of the trained GM to infer their current location based on the local RSS collected. The GM (copy) is then retrained on the local RSS to create local models (LMs), which are subsequently uploaded back to the server for aggregation with the GM [15], [16], [17].

Although FL enables distributed training across heterogeneous devices, it typically aims to make the GM parameters converge toward a fixed optimum. In the context of indoor localization, this implies that the GM learns a fixed mapping between RSS fingerprints and RPs such that, after multiple rounds of aggregation, subsequent LM updates result in only minor GM parameter adjustments. This assumption may reasonably hold when RSS variations are primarily induced by device heterogeneity, discussed in detail in Section 3. However, RSS fluctuations caused by temporal dynamics of the indoor environment continuously reshape the underlying data distribution in an unpredictable manner, discussed in detail in Section 3 [14]. This unpredictability can violate the assumption of convergence to a fixed optimum. In such settings, enforcing convergence is suboptimal. Instead, the GM must continuously adapt to evolving RSS patterns while preserving previously learned representations [18]. This scenario is more accurately characterized as FL operating under dynamic or continually evolving datasets, which we refer to as continual federated learning (CFL).

The CFL approach allows a long-term use case for the localization system by being continuously updated based on changes in indoor environment characteristics. However, the approach has not yet been explored for the mobile indoor localization problem domain. For indoor localization, the approach creates new challenges, making the localization system susceptible to model corruption due to the temporal environment dynamics of indoor spaces and from exposure to potential model poisoning attacks over extended durations of time [19], [20]. The introduction of temporal environment dynamics and model poisoning attacks can cause significant alterations to the learned weight tensors from the offline phase [21]. These alterations can lead the LM weight tensors to significantly deviate from the current state of the GM. This extremely deviated LM, also termed as corrupted LM, introduces corruption to the GM through the aggregation process [22]. The effects of this corruption are exacerbated over multiple rounds of LM updates over time where the GM integrates erroneous information from the LMs. This leads to degraded performance and abnormal behavior of the GM [22], [23]. Therefore, robust techniques are necessary to monitor and safeguard the GM from such issues, which is one of the key contributions of ARMOR.

Furthermore, the presence of model poisoning attacks can introduce deliberate corruptions to the LM with the intention to compromise the GM [24]. Therefore, it is crucial to detect such malicious activities and monitor the learning trajectory of the GM to prevent failure of the localization system. Monitoring this trajectory helps in identifying unusual patterns or deviations caused by either temporal environment dynamics or model poisoning attacks. In this paper, we introduce ARMOR, a novel framework that addresses the multifaceted challenges of device heterogeneity, temporal environment dynamics, model corruption and model poisoning attacks in a CFL-based solution for indoor localization. The novel contributions of this work are:

- To the best of our knowledge, ARMOR is the first CFL-based framework designed to address model corruption induced through device heterogeneity, temporal environment dynamics, and model poisoning attacks in indoor localization.
- ARMOR introduces a novel state-space model (SSM) that continuously monitors the GM's learning trajectory, predicts its expected evolution based on historical trends, and detects deviations indicative of corruption.
- ARMOR introduces a novel adaptive federated aggregation technique to safeguard the GM from model corruption.
- ARMOR is designed for lightweight deployment on mobile devices without compromising accuracy or privacy of the localization system.
- Evaluations are conducted across multiple real building environments, incorporating several months of temporal variations. We rigorously test the framework for resilience against different model poisoning attacks and perform benchmarking against state-of-the-art frameworks.

The rest of this draft is organized as follows. Section 2 discusses relevant prior related works. Section 3 discusses RSS fluctuations induced by mobile device heterogeneity and temporal environmental dynamics that contribute to model corruption. Section 4 examines various model poisoning attack methods that induce deliberate model corruption. Section 5 describes the ARMOR framework that is designed to mitigate model corruption. Section 6 presents the experimental results. Finally, Section 7 provides the conclusion.

## 2. RELATED WORK

Wi-Fi fingerprinting-based indoor localization has garnered significant attention, with premiere research conferences like IPIN (International Conference on Indoor Positioning and Indoor Navigation) [25] and industry giants like Microsoft [26] hosting competitions to advance this field. Traditional ML approaches, including K-Nearest Neighbors (KNN) [27], Hidden Markov Model (HMM) [11], Gaussian Process Classification (GPC) [28], and Extreme Gradient Boosted Trees (XGBoost) [53] have demonstrated some success in tackling the challenges of RSS fluctuations in indoor environments. These methods attempt to account for the variability introduced by factors like human movement, obstacles, and signal interference [29]. However, fully overcoming fluctuations induced by temporal environment dynamics remains an open challenge. The inherent unpredictability of indoor settings and the complex nature of RSS signal variations continue to pose significant hurdles for traditional ML models, preventing them from consistently achieving high localization accuracy.

Moreover, the issue of device heterogeneity compounds these challenges, introducing additional variability into the RSS fingerprinting process. This heterogeneity stems from differences in Wi-Fi chipsets and noise filtering software used by various manufacturers, which are critical for extracting RSS fingerprints. Such variability complicates the task for traditional ML-based indoor localization systems [29], [30]. To address these challenges, researchers have turned to deep learning (DL) algorithms, leading to the development of frameworks like BmmW [31], MLPLOC [32], CNNLOC [33], SANGRIA [34], ANVIL [35], TIPS [36] ,VITAL [37], GoPlaces [51], WiKAN [52], AATS [53]. BmmW [31] and MLPLOC [32] leverage deep neural networks (DNNs) with enhanced RSS pre-processing techniques to improve feature correlation. CNNLOC [33] introduces a modified convolutional neural network (CNN) to better capture relevant features in RSS fingerprints. SANGRIA [34] employs autoencoders based on DNNs, while ANVIL [35] uses attention neural networks to focus on critical input features. TIPS [36], VITAL [37], and AATS [53] utilize transformer-based encoding to improve resilience against fluctuations caused by device heterogeneity. GoPlaces [51] utilizes a privacy preserving attention-LSTM based model to detect movement of the mobile device indoors. WiKAN [52] improves heterogeneity resilience of DNN based solutions by using lightweight Kolmogorov–Arnold Networks (KANs). While these frameworks provide significant improvements in addressing the device heterogeneity challenge, they still struggle to actively learn and adapt to the temporal dynamics of indoor spaces, limiting their long-term use case.

Frameworks like SELE [38], STONE [39] and STELLAR [40] attempt to establish long-term prediction stability by using approaches such as Siamese neural networks and contrastive learning to estimate future changes in indoor environments. While these approaches are designed to predict long-term environmental changes, they are limited in their ability to adapt to the continuous fluctuations that occur in real-world indoor environments. This is because they rely heavily on static models that cannot dynamically incorporate new information from various devices distributed throughout the indoor environment. To effectively adapt to these ongoing changes, a more distributed ML approach is required — one that can leverage data collected from different devices across various indoor locations while ensuring user privacy [55]. Federated learning (FL) offers a promising solution by enabling multiple clients to collaboratively update a centralized global model (GM) without sharing raw data, thus preserving user privacy [41], [56]. FL-based systems like FedLoc [41] and FedHIL [42] have shown promise, with FedLoc [41] utilizing federated stochastic gradient descent (FedSGD) [43] to facilitate a privacy preserving distributed ML. However, this approach is highly susceptible to model corruption as erroneous updates from clients can skew the GM's learning trajectory, leading to abnormal behavior of the localization system. FedHIL [42] employs domain-specific selective weight aggregation strategies to mitigate the impact of erroneous client updates and maintain stability. However, it remains vulnerable to deliberate model poisoning attacks [57], [58].

A few frameworks like KRUM [45], Multi-KRUM [45], and Bulyan [46] aim to tackle model poisoning attacks in an FL setting. KRUM [45] performs Euclidean distance-based filtering to select the local model (LM) update that is least deviating from the majority of LM updates from other clients. Multi-KRUM [45] extends the KRUM [45] method by iteratively aggregating multiple of such LM updates. Bulyan [46] further filters out outliers, by using a combination of trimmed-mean and majority vote before applying federated averaging (FedAVG) [46]. Despite these efforts, the challenge of deploying FL-based indoor localization systems in a CFL manner exacerbates the potential for GM corruption, due to the continuous LM updates over extended durations of time. In this setting, there is a need for sophisticated mechanisms to not only mitigate corruption but also continuously monitor and detect potential corruptions from being aggregated with the GM.

Recognizing the limitations of prior work in addressing model corruption induced by device heterogeneity, temporal dynamics, and adversarial threats, we propose ARMOR, a novel CL–based framework that proactively safeguards the GM against model corruption. Unlike existing methods that either passively aggregate updates or treat heterogeneity, temporal dynamics and security separately, ARMOR introduces a state-space model (SSM) that continuously monitors the evolution of the GM weights, predicts its expected learning trajectory based on historical trends, and detects deviations caused by erroneous or malicious updates. Upon detecting such deviations, ARMOR employs a novel adaptive federated aggregation strategy that selectively regulates and filters incoming updates before they influence the GM. Through this proactive and trajectory-aware protection, ARMOR maintains robust feature representations in the GM and high localization accuracy despite model poisoning attempts and real-world temporal fluctuations.

## 3. BACKGROUND: WI-FI RSS FLUCTUATIONS IN INDOOR LOCALIZATION

Wi-Fi RSS measurements reflect the power level of the wireless signal received from a Wi-Fi AP by the mobile device. This measurement is expressed in decibels relative to a milliwatt (dBm) and typically ranges from -100dBm to 0dBm. The value -100dBm indicates no signal or an invisible AP and 0dBm represents the strongest signal [7]. In real-world scenarios, several factors can cause fluctuations in RSS values collected by the mobile device [8]. These fluctuations can be broadly categorized into two main types: device heterogeneity-induced fluctuations and temporal environment dynamics-induced fluctuations.

- Device Heterogeneity-Induced Fluctuations: These occur due to inherent differences in the hardware, software, and firmware of various mobile devices. Different manufacturers design their devices with custom Wi-Fi chipsets, leading to differences in antenna shape, size, location, and power gain. Even when devices use the same Wi-Fi chipset, they can still experience RSS fluctuations due to differences in signal processing methods used to filter and smooth RSS noise, as well as different data parsing algorithms, firmware, operating systems, and sensor calibrations, all of which can induce RSS fluctuations [35]. The resulting inconsistent RSS readings lead to errors in location estimation, making the localization system less accurate. A viable localization framework must be robust enough to handle device-specific fluctuations. FL presents a promising approach for addressing this challenge. By leveraging the collaborative learning process inherent in FL, indoor localization systems can learn and adapt to heterogeneity-induced fluctuations across various devices. However, addressing heterogeneity alone is not sufficient due to the presence of temporal dynamics-induced fluctuations.

- Temporal Dynamics-Induced Fluctuations: These refer to the fluctuations in RSS values that occur over time due to changes in the indoor environment. Such fluctuations can be caused by factors such as the movement of people, rearrangement of furniture, and changes in AP configurations, such as turning the APs on or off for maintenance purposes [40]. Temporal dynamics can introduce both short-term and long-term fluctuations in RSS values. Short-term fluctuations are often abrupt and significant, resulting from sudden changes like people walking by or objects being moved, which can cause notable deviations in the RSS measurements. These abrupt changes lead to large variations in the input features of client devices, causing the weights of the LMs to deviate significantly from those of the GM. When these deviated LMs are aggregated with the GM, they can corrupt the GM's learning trajectory, leading to weight patterns that no longer accurately represent the true dynamics of the indoor space, potentially causing the localization system to behave abnormally or fail. Additionally, temporal dynamics can also be impacted by long-term changes in the indoor environment, such as permanent furniture rearrangements or alterations in AP configurations such as replacing old APs with newer ones. Unlike short-term fluctuations, these long-term changes are important for the localization system to learn and adapt to, rather than filter out. A robust solution must be capable of distinguishing between these short-term and long-term fluctuations, allowing it to adapt to genuine changes in the environment while mitigating the impact of transient noise.

To illustrate the fluctuating nature of RSS, we visualize the RSS values collected across different commercially available mobile devices at the same RP (i.e., location) but over different time instances in Figure 1. This data is obtained from the open-source CSUIndoorLoc [50] dataset (discussed in more detail in Section 6.2). In Figure 1, the X-axis represents the MAC (Media Access Control) address of a subset of Wi-Fi APs, while the Y-axis represents the RSS values captured by six different mobile devices, measured in dBm. Each box plot in the figure includes upper and lower whiskers, and average (yellow horizontal line in boxes), which display the maximum, minimum, and average RSS values captured by the mobile devices at the same RP and time respectively. Boxes of the same color represent RSS values captured at the same RP at a specific instance of time, with the five unique colors representing data captured at day 1, then at day 7, at day 30, at month 4, and at month 8. The gap between the upper and lower whiskers in the RSS values for each AP reflects the extent of fluctuations captured across different mobile devices (heterogeneity-induced fluctuations).

We can observe device heterogeneity-induced fluctuations by looking at the box-whisker instances for the same time instance in Figure 1. For example, at day 30, the most significant heterogeneity-induced fluctuation is observed at the MAC address "80:8d:b7:55:35:12", with RSS values ranging from -70 dBm to -90 dBm (fluctuation of -20 dBm). In contrast, smaller differences are observed at "38:17:c3:1f:0e:01" on day 1, with values ranging from -71 dBm to -76 dBm (fluctuation of -5 dBm).

Some cases, such as at "38:17:c3:1f:0e:11", show minimal differences between the upper and lower whiskers across most time intervals, indicating the absence of heterogeneity-induced fluctuations. Given the smaller magnitude of these heterogeneity-induced fluctuations, they result in a less or non-deviated LM, thereby reducing the chances of GM corruption.

Analyzing the temporal environment dynamics aspect of the same data reveals larger fluctuations, with significant changes in RSS observed across different intervals of time. For instance, at "ea:9e:b4:53:24:54", the highest RSS is recorded on day 30 with -60 dBm and the lowest RSS is recorded on month 8 with -100 dBm (fluctuation of -40 dBm). Similar trends are observed across all other MAC addresses (each of which identifies a unique AP). These fluctuations are much more pronounced compared to device heterogeneity-induced variations and upon aggregation with the GM, can alter the learning trajectory of the GM, leading to corruption of the GM. The RSS fluctuations are mainly introduced due to RSS being altered either by the mobile device's hardware or software configuration leading to the device heterogeneity effect. Furthermore, temporal dynamics such as Wi-Fi signal reflection from furniture, human movement, and other electronics equipment can cause more significant changes in RSS, as shown in figure 1. These RSS during the online phase could be different from the offline phase causing the underlying ML model to mis predict locations indoors.

In a CFL setting, a stable learning trajectory ensures that the GM adapts to evolving data while retaining past knowledge and not converge to a fixed optimum, which is crucial for maintaining accuracy amid RSS fluctuations. Deviations from this trajectory can lead to model corruption, reduced accuracy, and system failure. A robust GM monitoring and aggregation technique is essential to mitigate such GM corruption.

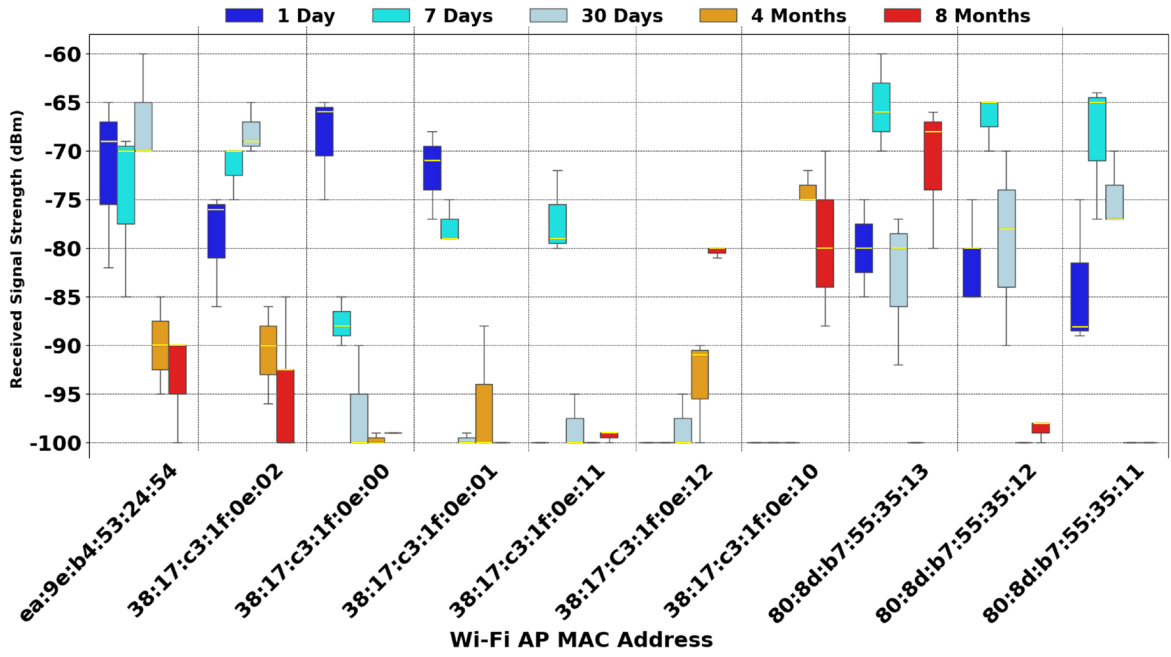

Figure 1. Analysis of RSS fluctuation in indoor localization testing presence of device heterogeneity and temporal dynamics induced fluctuations at a specific RP (indoor location).

## 4. MODEL POISONING ATTACKS IN INDOOR LOCALIZATION

Model poisoning attacks involve deliberately introducing corruptions to the GM with the intent to degrade the GM's accuracy and induce abnormal behavior, ultimately leading to the failure of the localization system [44]. These attacks occur when an adversary maliciously injects corruptions into the GM, thereby altering its learning trajectory. In CFL, model poisoning attacks are executed during the client's LM updates sent to the server, as discussed next.

Figure 2 shows the workings of a CFL-based indoor localization system and how model poisoning attacks are possible within it. In this system, the server maintains a GM, which is made available to all clients within the indoor space. Each client downloads a copy of the GM, stores it as the LM, and uses it to provide location predictions based on the local RSS fingerprints captured. Over time, the LM is retrained on the local RSS data captured by a client (local training), updating the LM with the local dynamics of the indoor space in the vicinity of the client. The updated LM is then sent back to the server, which aggregates the weight tensors from all LMs to update the GM. This approach ensures privacy of the client's local data, as locally captured RSS data remains within the client device.

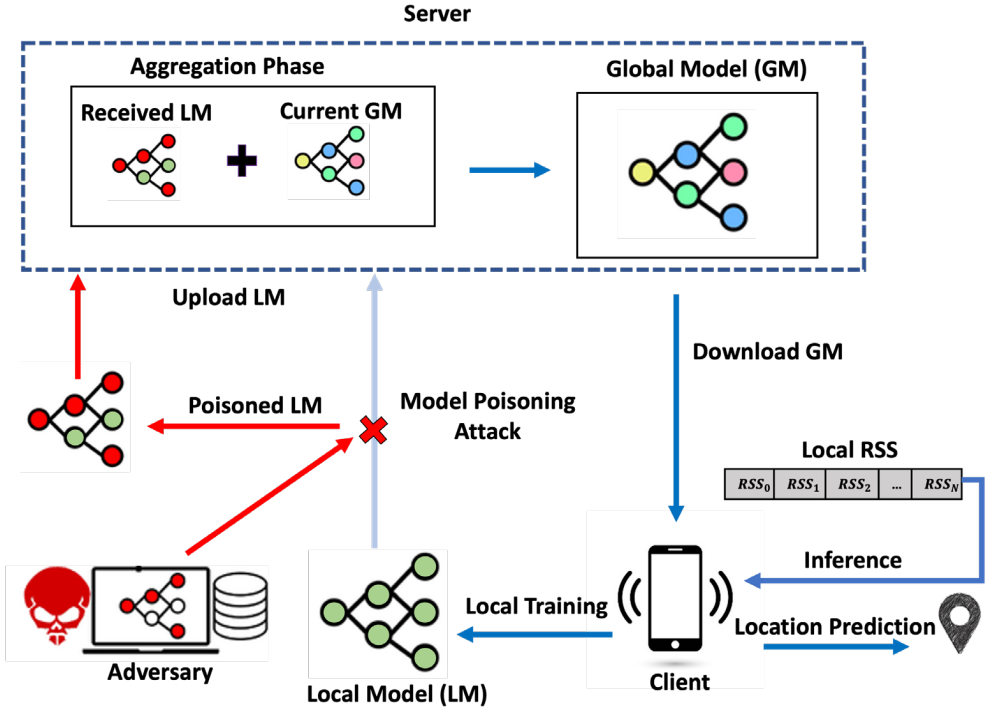

Figure 2. Execution of model poisoning attack in CFL based indoor localization.

Adversaries can introduce corruptions to the LM weight tensors before the aggregation phase in the server. In a non-attack scenario, only a subset of the model's weights or certain tensors may experience significant changes due to RSS fluctuations, particularly those tensors more directly tied to the input features affected by these fluctuations. However, an attacker can modify the LM weights to cause widespread changes across many tensors, inducing model corruption. Successful model poisoning attacks aim to remain covert as introducing arbitrary corruptions can be easily detected by anomaly detection and intrusion detection systems [46], [47], [48]. The impact of model poisoning attacks is exacerbated in a CFL setting, where one or several devices continuously update corrupted LMs back to the server for aggregation.

### 4.1 Model Poisoning Attack Methods

In a CFL setting, adversaries can leverage various model poisoning attack methods to compromise the GM. These attacks can target specific vulnerabilities in the system, allowing the adversary to influence the localization accuracy or even cause system-wide failure. The following are some popular attack strategies that can disrupt the GM's learning process, tailored for indoor localization:

- **Random Attack:** The adversary generates arbitrary updates to the LMs without any specific pattern or consistency. The goal is to introduce noise and random perturbations into the GM's aggregation process. This method disrupts the GM learning trajectory by introducing high variability and unpredictability into the updates, as highlighted by equations 1 and 2:

$$\Delta W'_T = \Delta W_T + N_T \qquad (1)$$

$$W_{T+1} = W_T + \frac{1}{K}\sum_{i=1}^{K} \Delta W'_{T,i} \qquad (2)$$

where $W_T$ represents the weight tensors of the GM at time $T$. The legitimate LM weight tensors $\Delta W_T$ are derived from local data. $N_T$ denotes the perturbated tensors to be added to $\Delta W_T$, resulting in the random attacked weight tensors $\Delta W'_T$ in equation 1. $\Delta W'_T$ is aggregated with the current GM weight tensors $W_T$ using equation 2, where $K$ is the number of clients.

- **Gaussian Attack:** Here, the adversary injects malicious updates to the LM that are sampled from a Gaussian distribution such as a probability density function $\Psi$. The goal of this method is to introduce statistically consistent noise into the GM's aggregation process, which can subtly alter the learning trajectory of the GM without introducing obvious outliers that can be easily detected. This type of attack can be particularly insidious as the updates may appear legitimate due to their statistical properties, and can be expressed by:

$$\Psi_T = \frac{1}{\sigma\sqrt{2\pi}}\, e^{-\frac{(x-\mu)^2}{2\sigma^2}} \qquad (3)$$

$$\Delta W'_T = \Delta W_T + \Psi_T \qquad (4)$$

where $x$ denotes the attack strength, $\mu$ is the mean of the Gaussian distribution, $\sigma$ is the standard deviation, $\sigma^2$ is the variance, and $e$ represents the base of natural logarithm. This results in the Gaussian attacked weight tensors $\Psi_T$ at time $T$ from equation 3. $\Psi_T$ is then added to the legitimate LM weight tensors $\Delta W_T$ from equation 4.

- **History Attack:** Here, the adversary exploits historical data to generate malicious updates intended to corrupt the GM's learning trajectory. The attacker analyzes past LM updates to generate new updates that appear legitimate but are strategically designed to corrupt the GM. The adversary collects historical GM weight tensors $\{W_{T-1}, W_{T-2}, \ldots, W_{T-J}\}$, where $T$ is the current time, and $J$ representing the number of past updates considered. By analyzing these updates, the adversary identifies patterns and trends in the model's evolution, as highlighted by:

$$\Delta W'_T = \Delta W_T + \beta \cdot f(\{W_{T-1}, W_{T-2}, \ldots, W_{T-J}\}) \qquad (5)$$

where $\Delta W_T$ is the legitimate LM update at time $T$, $\beta$ denotes the scaling factor used to amplify the perturbation strength, and $f$ is the function that aggregates historical updates. This results in the history attacked weight tensors $\Delta W'_T$.

- **Model Poisoning Federated (MPAF) Attack:** Here, the adversary utilizes surrogate models to generate malicious updates introduced through fake clients [49]. The adversary first trains surrogate models to mimic the GM's behavior, using them to generate poisoned updates. These updates, which include a malicious component, are submitted by fake clients that appear legitimate, as highlighted by

$$\Delta W'_T = \Delta W_T + \beta \cdot S_T \qquad (6)$$

where $\Delta W_T$ is the legitimate LM update at time $T$, $\beta$ denotes the scaling factor used to amplify the perturbation strength, and $S_T$ denotes the tensors generated by the surrogate model to be added to $\Delta W_T$, resulting in the MPAF attacked weight tensors $\Delta W'_T$.

*4.2 Formulation of Model Poisoning Attacks*

In formulating model poisoning attacks for indoor localization systems, we employ the four distinctive methods discussed above: Random, Gaussian, History, and MPAF attacks. Our objective is to generate model poisoning attacks by introducing corruptions that modify the LM tensors in a way to induce corruption in the GM within the CFL setting. To effectively simulate potential real-world model poisoning attacks, we leverage two key parameters:

- **Percentage of Tensors Attacked (*ATT*):** We assess the susceptibility of CFL based indoor localization systems to various potent model poisoning attacks, such as Random $N_T(ATT)$, Gaussian $\Psi_T(ATT)$, History $\beta \cdot f(\{W_{T-1}, W_{T-2}, \ldots, W_{T-J}\}(ATT))$, and MPAF $\beta \cdot S_T(ATT)$ attacks to generate poisoned or corrupted LMs that are

8uploaded to the server for aggregation with the GM. One crucial parameter that an adversary can control is *ATT* of the legitimate LM tensors that can be replaced by the adversary to induce corruption. The adversary typically aims to remain covert while uploading corruptions to the server to avoid detection by anomaly detection and intrusion detection systems. We introduce the *ATT* parameter, which ranges from 0% to 100%. This parameter represents the proportion of LM tensors that are attacked: 0% indicates that none of the legitimate LM tensors have been replaced or corrupted by the poisoning methods and 100% indicates that all legitimate LM tensors have been replaced or corrupted by the adversary. By varying the *ATT* parameter, we can systematically study the impact of different levels of corruption on the learning trajectory of the GM.

- **Learning Trajectory of the GM:** The learning trajectory refers to the path that the GM parameters follow during the CFL process. In the context of CFL, it is crucial to monitor the learning trajectory to ensure that the GM continuously adapts to new data without forgetting previously learned information. The learning trajectory provides insights into the stability and convergence of the GM, especially when subjected to model corruptions. To quantify the learning trajectory, we employ the Frobenius norm, a normalization method which measures the difference between the current GM and the updates received from LMs. The Frobenius norm is a matrix norm that is particularly useful for comparing GM-LM parameters in the CFL setting.

$$||A||_F = \sqrt{\sum_{i=1}^{m} \sum_{j=1}^{n} |a_{ij}|^2} \qquad (7)$$

$$Frobenius\ norm = ||W_T - \frac{1}{K}\sum_{i=1}^{K} \Delta W_{T,i}||_F \qquad (8)$$

In equation 7, $A$ is a matrix with elements $a_{ij}$ and $||A||_F$ represents the Frobenius norm of matrix $A$. In terms of detecting model corruption, we use the Frobenius norm to calculate the deviation of the GM after aggregation of local updates, indicating the extent of corruption. In equation 8, $W_T$ represents the weight tensors of the GM at time $T$, $\Delta W_{T,i}$ represents the update from the $i^{th}$ LM, and $K$ is the number of clients.

In summary, we explore various model poisoning methods that can induce significant corruption in CFL-based indoor localization systems, compromising the underlying GM. This underscores the need for a robust corruption monitoring system that can continuously monitor and safeguard the GM against potential threats. The ARMOR framework is designed to address these challenges with advanced techniques to monitor, detect, and mitigate model corruption within the CFL process. The details of the ARMOR framework are discussed in the next section.

## 5. ARMOR FRAMEWORK

The ARMOR framework is designed to enhance the reliability of indoor localization systems within the CFL setting by detecting and mitigating model corruption introduced through erroneous LM updates. ARMOR achieves this by monitoring and safeguarding the evolution of the GM at the tensor level, preserving its optimal learning trajectory. ARMOR consists of three core components: (i) the corruption monitoring phase, that utilizes a novel state-space model (SSM) to track the GM's learning trajectory and predict its next weight state prior to LM aggregation, (ii) the corruption detection phase, which detects deviations between LM updates and the SSM's projected weight state that could corrupt the GM's learning trajectory, and (iii) the corruption mitigation phase, which applies a novel adaptive federated aggregation mechanism to selectively regulate and filter LM updates, thereby preventing corruption from propagating into the GM.

As illustrated in Figure 3 (a) and algorithm 1, the framework begins in the offline phase, where RSS fingerprints are captured across different RPs (i.e., locations) within a building floorplan. Multiple fingerprints are collected per RP to effectively capture data variability. These fingerprints are labeled and stored in an RSS fingerprint database, forming the offline training data for the ARMOR framework. The offline training data is then divided into two parts to train the GM and SSM sequentially (line 1 in algorithm 1), as shown in steps 1 and 2 in Figure 3 (a). The first part of the offline training data is used to train the GM alone (steps 3 and 4 in Figure 3 (a) and line 2 in algorithm 1). After training the GM, it is retrained on the second part of the offline training data (step 5 in Figure 3 (a) and lines 3 to 8 in algorithm 1). During this retraining, we explicitly record the GM's weight tensors after each retraining class (step 6 in Figure 3 (a) and lines 4 to 7 in algorithm 1), thereby constructing a historical sequence of weight states ($W_0, W_1, \ldots, W_{T-1}$). Here $W_T$ denotes the GM weight tensors at retraining time $T$. This GM history data is used as training input for the SSM (step 7 in Figure 3 (a) and line 8 in algorithm 1), allowing the SSM to learn the learning trajectory of the GM after initial training. The SSM is then used to generate a projection of the next expected GM weights ($P_{T+1}$) (step 8 in Figure 3 (a)), which is part of the corruption monitoring phase. This projected weight serves as a reference for what the next GM weights ($W_{T+1}$) are expected to achieve based on the GM's historical updates, as shown in algorithm 2.

In the online phase, a client within the indoor space captures local RSS fingerprints and receives a copy of the GM from the server, as shown in step 1 in Figure 3 (b) (lines 1 to 3 in algorithm 2). The client uses the GM to infer location coordinates based on the local data (step 2 in Figure 3 (b)). The GM copy then performs local training on the local RSS data, creating the LM ($\Delta W_{T+1}$) (steps 3 and 4 in Figure 3 (b) and line 2 in algorithm 2). The LM is then uploaded to the server for aggregation. At the server side, the SSM continuously monitors the evolution of GM weights (learning trajectory) and predicts the next expected set of GM weights based on historical trends ($P_{T+1}$) (Steps 5 and 6 in Figure 3(b) and line 4 in algorithm 2). Upon receipt of $\Delta W_{T+1}$ and $P_{T+1}$, the server initiates the corruption detection phase, in which the received LM weights are compared against the projected GM weights predicted by the SSM (Steps 7 and 8 in Figure 3(b)). This corruption detection phase is used to detect deviations between the SSM projected weights with the received LM weights at the tensor level that may corrupt the GM, as detailed in Section 5.2. Subsequently, a corruption mitigation phase is activated to selectively attenuate the deviated tensors identified between $\Delta W_{T+1}$ and $P_{T+1}$, as detailed in Section 5.3 (Step 9 in Figure 3(b) and lines 5 to 8 in algorithm 2). The resulting sanitized update, denoted as $\Delta W'_{T+1}$ is then forwarded to the aggregation phase (Step 10 in Figure 3(b) and line 9 in algorithm 2), where only reliable tensors are aggregated into the current GM, thereby mitigating the propagation of corruption. The updated GM then updates the SSM, guiding it to provide effective projections for the next round of LM updates $P_{T+2}$ (steps 11 and 12 in Figure 3(b) and line 10 in algorithm 2).

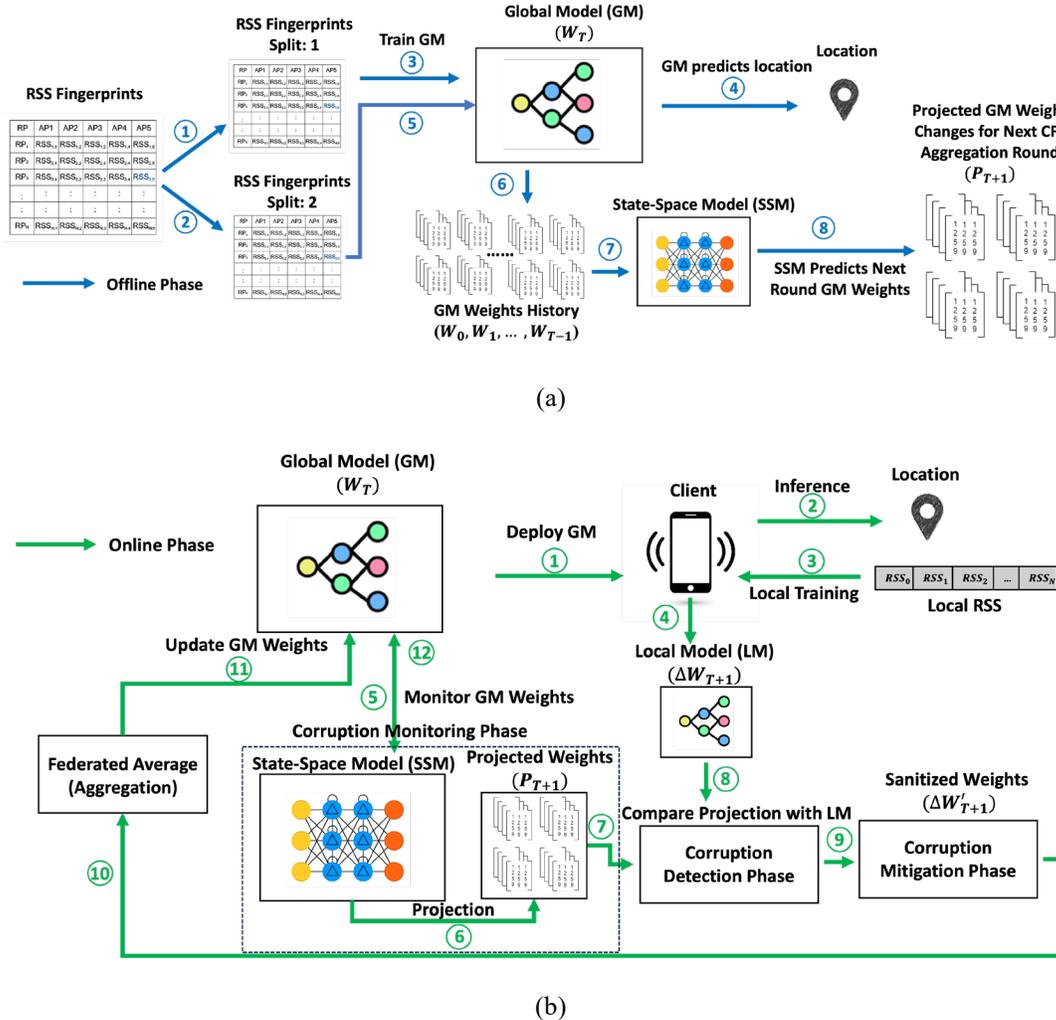

Figure 3. Working of the ARMOR framework (a) ARMOR offline phase, (b) ARMOR online phase

In summary, the ARMOR framework offers a promising approach to ensuring the integrity of the GM within the CFL setting for indoor localization. By capturing and leveraging historical GM weight data, implementing robust corruption detection mechanisms, and preventing the aggregation of corrupted updates, ARMOR enhances the resilience of the system against model corruptions induced by RSS fluctuations and model poisoning attacks. This framework ensures continuous learning and accurate localization even in the presence of adversarial threats, thereby maintaining the reliability and performance of the system. In the following sections, we discuss the workings of the corruption monitoring phase, corruption detection phase, and corruption mitigation phase, explaining how each contributes to safeguarding the GM.

---

**ALGORITHM 1:** ARMOR Offline Phase

---

**Input:** Offline training RSS fingerprints $Data_{Offline}$
**Output:** Trained Global Model (GM) $GM_T$; Trained State-Space Model (SSM) $SSM_T$
1: $Data_{GM}, Data_{SSM} \leftarrow$ Split ($Data_{Offline}$)
2: $GM_T \leftarrow$ Train GM ($Data_{GM}$)
3: $History_{GM} \leftarrow \{\}$
4: **For** each re-training $Class_{RP} \in Data_{SSM}$ **do**
5:    $GM_T \leftarrow$ Re-Train GM ($Class_{RP}$)
6:    $History_{GM} \leftarrow$ Append $GM_T$
7: **End**
8: $SSM_T \leftarrow$ Train SSM ($History_{GM}$)
9: Return $GM_T, SSM_T$

---

**ALGORITHM 2:** ARMOR Online Phase

---

**Input:** Trained GM $GM_T$; Trained SSM $SSM_T$; Client set $\{c_0, c_1, \ldots c_k\}$
**Output:** Updated GM $GM_{T+1}$; Updated SSM $SSM_{T+1}$
1: **For** each client $c_k \in \{c_0, c_1, \ldots c_k\}$ **do**
2:    $\Delta GM_k \leftarrow$ Local Train($GM_T, Data_k$)
3: **End**
4: $P_{T+1} \leftarrow$ Corruption Monitoring ($SSM_T, GM_T$)
5: **For** each received $\Delta GM_k$ **Do**
6:    $S_k \leftarrow$ Corruption Detection ($\Delta GM_k, P_{T+1}$)
7:    $\Delta GM'_k \leftarrow$ Corruption Mitigation ($\Delta GM_k, P_{T+1}$)
8: **End**
9: $GM_{T+1} \leftarrow GM_T +$ Average ($\Delta GM'_k$)
10: $SSM_{T+1} \leftarrow$ Update ($SSM_T, GM_{T+1}$)
11: Return $SSM_{T+1}, GM_{T+1}$

---

**ALGORITHM 3:** ARMOR Corruption Monitoring, Detection and Mitigation Phases

---

**Input:** Trained GM $GM_T$; Trained SSM $SSM_T$; Client set $\{c_0, c_1, \ldots c_k\}$
**Output:** Updated GM $GM_{T+1}$; Updated SSM $SSM_{T+1}$
1: $P_{T+1} \leftarrow SSM_T (GM_0, GM_1, \ldots, GM_T)$
2: **For** each client update $\Delta GM_{T+1}^k$ **do**
3:    $S_k \leftarrow Cosine (\Delta GM_{T+1}^k, P_{T+1})$
4:    **if** $S_k > 0$ **then**
5:       $\Delta GM_{T+1}^k \leftarrow \Delta GM_{T+1}^k$
6:    **else**:
7:       **For** each tensor $i$ **do**
8:          $D_i \leftarrow \Delta GM_{T+1,i}^k - P_{T+1,i}$
9:       end

10:        $\phi \leftarrow \frac{1}{k}\sum_{i=1}^{k}|D_i|$
11:     **For** each tensor $i$ **do**
12:        $\Delta GM_{T+1,i}^{k} \leftarrow \Delta GM_{T+1,i} - \phi$
13:     **end**
14:   **end**
15: **end**
16: $GM_{T+1} \leftarrow GM_T + \frac{1}{k}\sum_{i=1}^{k}\Delta GM_{T+1,i}^{k}$
17: $SSM_{T+1} \leftarrow$ Update $(SSM_T, GM_{T+1})$
18: Return $GM_{T+1}, SSM_{T+1}$

### 5.1 Corruption Monitoring Phase

The corruption monitoring phase is crucial not only to monitor the learning trajectory of the GM but also to continuously evolve and adapt to the learning trajectory, as shown in algorithm 3. This phase is designed to detect any anomalies in the GM that may indicate corruption, thereby ensuring the integrity and reliability of the GM over time, as shown in steps 5, 6, and 7 in Figure 3(b) and line 1 in algorithm 3. The SSM serves as the core of this corruption monitoring phase, with its primary component being a modified Gated Recurrent Unit (GRU) network. This GRU network is trained on the history of the GM weight tensors, as illustrated in Figure 4. The GRU learns the differences between the tensors in the GM history, allowing it to understand the typical patterns of change in the GM weight tensors. By capturing these patterns, the GRU can predict the next set of GM weight tensors based on the learned patterns of differences. Next, we provide a detailed explanation of the SSM architecture utilized in the ARMOR framework.

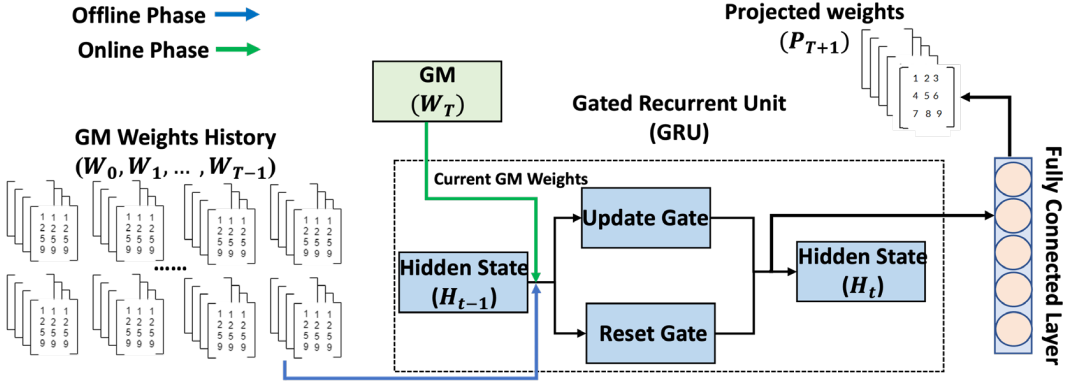

Figure 4. SSM architecture used in the corruption monitoring phase within the ARMOR framework.

**SSM Architecture:** The SSM in the ARMOR framework consists of a bidirectional GRU layer followed by a fully connected layer. This architecture is designed to capture sequential patterns in the GM history weight tensors and predict the next set of GM weight tensors. In this setup, the GRU layer processes GM weight tensors and maintains memory, which is crucial for learning the changes in GM weight tensors over time. The GRU layer consists of several GRU cells and each GRU cell contains two primary gates: the reset gate and the update gate, which control the flow of information. The reset gate denoted as $R_T$, determines how much of the past information should be forgotten, as expressed by:

$$R_T = Sigmoid(W_R \cdot [H_{T-1}, X_T] + B_R) \quad (9)$$

where $W_R$ is the weight matrix associated with the reset gate, $B_R$ is the bias tensor, and $Sigmoid$ ensures that gate values range between 0 and 1. The update gate, $Z_T$ controls how much of the past information needs to be retained, as expressed by:

$$Z_T = Sigmoid(W_Z \cdot [H_{T-1}, X_T] + B_Z) \qquad (10)$$

where, $W_Z$ and $B_Z$ are the weight matrix and bias tensors for the update gate, respectively. This gate decides the balance between retaining the past hidden state and updating it with the new candidate hidden state. The candidate hidden state, $H'_T$ represents the new memory content that is created by applying the reset gate to the previous hidden state, as expressed by:

$$H'_T = tanh\ (W_H \cdot [R_T \odot H_{T-1}, X_T] + B_H) \qquad (11)$$

where $W_H$ is the weight matrix for generating the candidate hidden state and $B_H$ is the bias tensor. Finally, the output of the GRU cell, the final hidden state $H_T$ is computed as:

$$H_T = (1 - Z_T) \odot H_{T-1} + Z_T \odot H'_T \qquad (12)$$

This equation combines $H_{T-1}$ and $H'_T$, modulated by the $Z_T$, where $\odot$ denotes element-wise multiplication. This combination allows the network to maintain relevant information while updating its memory with new information. After processing through the bidirectional GRU layer, the output is fed into a fully connected layer to generate the next set of weight predictions. The fully connected layer ensures that the output projections, denoted as $P_{T+1}$ match the dimensionality of the input, thereby maintaining consistency and accuracy in the predicted weights, as expressed by:

$$P_{T+1} = W_P \odot H_T + B_P \qquad (13)$$

where, $W_P$ is the weight matrix of the fully connected layer and $B_P$ is the bias tensors . This structure enables the SSM to provide precise projections of future weights, which are essential for detecting any deviations from the expected learning trajectory.

*5.2 Corruption Detection Phase*

The corruption detection phase from steps 7 and 8 in Figure 3 (b) is crucial for identifying LM tensors in the received LM weight tensors ($\Delta W_{T+1}$) that significantly deviate from the projected weight tensors ($P_{T+1}$) generated by the SSM, as illustrated in Figure 5 and algorithm 3. This phase ensures that any anomalies indicating potential corruption are promptly detected before the LM updates are aggregated into the GM, thereby maintaining the integrity and reliability of the GM. We use cosine similarity to detect these deviations (line 3 in algorithm 3). Cosine similarity measures the cosine of the angle between two non-zero tensors in a multi-dimensional space, providing a metric for how similar the two tensors are. Unlike existing techniques that rely solely on distance-based filtering, ARMOR leverages cosine similarity to detect fine-grained deviations from GM projections (generated by the SSM) rather than directly comparing GM weights. This enables the GM to be updated based on projected estimations of its next state, improving robustness against adversarial perturbations or model corruptions.

$$Cosine\ (\vec{a}, \vec{b}) = \frac{\vec{a} \cdot \vec{b}}{||\vec{a}|| \cdot ||\vec{b}||} \qquad (14)$$

In equation 14, $\vec{a} \cdot \vec{b}$ is the dot product of tensors $\vec{a}$ and $\vec{b}$. $||\vec{a}||$ is the Euclidean norm (magnitude) of tensors $\vec{a}$, and $||\vec{b}||$ is the Euclidean norm (magnitude) of tensors $\vec{b}$. In our framework, $\vec{a}$ represents $\Delta W_{T+1}$ and $\vec{b}$ represents $P_{T+1}$. The cosine similarity between these tensors are used tensors determine how closely the received updates align with the expected updates. If the cosine similarity value is greater than or equal to 0 (positive), it indicates that the tensors are closely aligned, suggesting that the received LM updates are likely legitimate. Conversely, if the cosine similarity value is lesser than 0 (negative), it indicates a deviation, suggesting potential corruption.

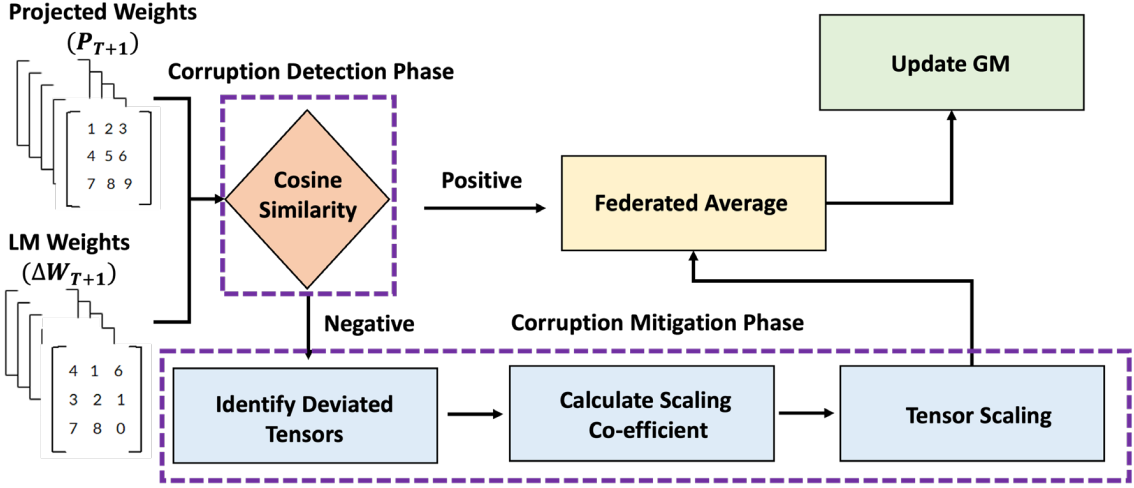

Figure 5. Corruption detection and mitigation phases in ARMOR.

*5.3 Corruption mitigation phase*

The corruption mitigation phase is the last phase in the ARMOR framework, designed to safeguard the GM from potentially corrupted LM updates, as shown in step 9 in Figure 3(b) and algorithm 3. Building on the corruption detection phase, where we use cosine similarity to flag potential corruptions, this phase ensures that only clean and reliable updates are aggregated into the GM. If the cosine similarity between the received LM updates ($\Delta W_{T+1}$) and the projected weights ($P_{T+1}$) is positive, it indicates that the updates are likely legitimate (line 4 and 5 in algorithm 3). In this case, we proceed with federated averaging (equation 18), averaging the current GM with the received LM updates (line 16 in algorithm 3). However, if the cosine similarity is negative, it suggests potential corruption (line 6 in algorithm 3). In such cases, we undertake a detailed process to mitigate the impact of the corrupted updates, as illustrated in Figure 5.

First, we identify the tensors in the LM updates that deviate significantly from the projected weights (lines 7 to 9 in algorithm 3), as given by:

$$D_i = \Delta W_{T+1,i} - P_{T+1,i} \quad (15)$$

where, $D_i$ represents the deviation of the $i^{th}$ tensor in the LM update ($\Delta W_{T+1,i}$) from its corresponding projected weight ($P_{T+1,i}$). By identifying these deviations, we can quantify how much each tensor in the received LM update differs from the expected value. Next, we compute a scaling coefficient ($\phi$) based on the observed deviations (line 10 in algorithm 3). This coefficient is calculated as the average deviation value of all tensors in the LM with respect to the projections, as indicated by:

$$\phi = \frac{1}{N} \sum_{i=1}^{N} |D_i| \quad (16)$$

where, $N$ is the total number of tensors, and $\phi$ represents the mean absolute deviation across all tensors. This scaling coefficient helps in adjusting the deviations uniformly across the LM update. Once we obtain $\phi$, we adjust $D_i$ in the LM by subtracting the $\phi$ value, ensuring that the final scaled LM tensors ($\Delta W'_{T+1,i}$) closely resemble the projections (lines 11 to 13 in algorithm 3), as shown by:

$$\Delta W'_{T+1,i} = \Delta W_{T+1,i} - \phi \quad (17)$$

In the next step, $\Delta W'_{T+1,i}$ represents the scaled LM tensor. This adjustment process brings the deviated tensors closer to their expected values, thus reducing the impact of corruption. After this adjustment, we perform federated averaging on the scaled LM with the current GM (lines 16 and 17 in algorithm 3), using the equation below:

$$W_{T+1} = W_T + \frac{1}{K}\sum_{i=1}^{K} \Delta W'_{T+1,i} \quad (18)$$

where, $W_{T+1}$ represents the updated GM, $W_T$ is the current GM, $K$ is the total number of clients, and $\Delta W'_{T+1,i}$ are the scaled LM updates from all clients. This federated averaging step ensures that the final aggregated model incorporates only the reliable and adjusted updates, maintaining the integrity and performance of the GM. By following this entire process, the ARMOR framework can mitigate the impact of corrupted LM updates, ensuring that the GM remains robust and reliable even in the presence of adversarial attempts to corrupt the GM.

## 6. EXPERIMENTS

### 6.1 Experimental Setup

In this section, we describe our experimental setup, designed to evaluate the performance of the proposed ARMOR framework in real-world scenarios and compare it with multiple state-of-the-art frameworks: KRUM [45], Multi-KRUM [45], Bulyan [46], FedHIL [42], and FedLoc [41]. Our experiments test the frameworks for resilience against model corruptions induced by several factors, including device heterogeneity, temporal variations, and model poisoning attacks. We utilize real-world RSS fingerprint data obtained from the open-source CSUIndoorLoc [50] dataset, discussed in the next subsection. To simulate real-world conditions, we train all the frameworks using data collected from a single device across multiple RPs in the building floorplan at the same collection instances (CIs) or time period. This setup enables the frameworks to be evaluated on heterogenous devices over extended periods (multiple CIs).

In the ARMOR framework, a modified Deep Neural Network (DNN) is used as the GM with 1 input layer, 1 output layer, and 2 hidden layers. Dropout and Gaussian noise layers are added to prevent the GM from overfitting. The input layer has 350 neurons, the hidden layers have 256 and 128 neurons respectively, and the output layer has a number of neurons equal to the number of RP classes. The dropout and Gaussian noise rates are set to 0.15. The Adam optimizer is used with a learning rate of 0.001, utilizing the sparse categorical cross-entropy loss function. The ReLU activation function is applied to each hidden layer, while the classification layer uses the Softmax activation function. The model is trained for 1,000 epochs, resulting in a total of 187,044 parameters, which corresponds to a compact model size of 730.64 KB. During the local training phase, we set the re-training epoch to 5 to ensure low overhead training on the mobile device.

The SSM consists of 1 bidirectional GRU layer containing 4 GRU cells and 1 fully connected output layer. We set the input and output dimensions for the SSM to be the same, allowing projections to have the same dimensions as the input. ReLU is used as the activation function in the GRU layer, and the linear activation function is used in the fully connected layer. The Adam optimizer is used with a learning rate of 0.01, employing the Mean Squared Error (MSE) loss function. The SSM is trained over 20 epochs, resulting in a total of 935,229 trainable parameters yielding a model size of 3.57 MB.

### 6.2 CSUIndoorLoc Dataset

The *CSUIndoorLoc* [50] is an open-source dataset released by the authors in 2022, consisting of Wi-Fi RSS fingerprints collected using publicly available mobile devices and building floorplans during regular working hours. Six mobile devices from different manufacturers, each equipped with a distinct Wi-Fi chipset, were used to collect the data. Table 1 provides an overview of the mobile devices utilized. It is worth noting that although some of the Wi-Fi chipsets used are the same, each mobile device manufacturer implements different firmware for noise filtering, resulting in inherent heterogeneity even when the Wi-Fi chipsets are identical. These mobile devices were selected to capture both heterogeneity due to hardware and software.

TABLE 1: DETAILS OF MOBILE DEVICES USED IN *CSUINDOORLOC*.

| Manufacturer | Model | Acronym | Wi-Fi Chipset | Operating System Version |
|---|---|---|---|---|
| BLU | Vivo 8 | BLU | MediaTek Helio P10 | Android 7.0 |
| HTC | U11 | HTC | Qualcomm Snapdragon 835 | Android 9.0 |
| Samsung | Galaxy S7 | S7 | Qualcomm Snapdragon 820 | Android 8.0 |
| LG | V20 | LG | Qualcomm Snapdragon 820 | Android 9.0 |
| Motorola | Z2 | MOTO | Qualcomm Snapdragon 835 | Android 9.0 |
| OnePlus | 3 | OP3 | Qualcomm Snapdragon 820 | Android 9.0 |

The RSS fingerprints were collected across six mobile devices in two distinct building floorplans, each with unique characteristics, to evaluate the indoor localization frameworks. Building 1 is made of a combination of wood and cement materials, has spacious areas with computers, and has a total of 160 identifiable Wi-Fi APs across a path length of 60 meters, resulting in 60 RPs (1 RP = 1 Meter). Building 2 has a more modern design with a mix of metal and wooden structures, open spaces, and shelves equipped with heavy metallic hardware. We identified 218 unique APs in Building 2 with a path length of 48 meters, resulting in 48 RPs. The dataset comprises a total of six fingerprints per RP, per device, per building floorplan at each CI, spanning a duration from 30 seconds to several months. A total of 10 CIs (CIs:0-9) were collected over a total span of 8 months. To capture the impact of varying human activity throughout the day, the first three CIs (0-2) for both paths were collected on the same day, with a time interval of 6 hours between each CI. This allowed for the collection of fingerprints early in the morning (8 A.M.), at mid-day (3 P.M.), and late at night (9 P.M.). Subsequently, CI 3 was collected after 24 hours, CI 4 was collected at day 7, CI 5 at day 30, CI 6 at month 2, CI 7 at month 3, CI 8 at month 4, and CI 9 at month 8.

For the Building 1 and Building 2 paths, we specifically utilized a subset of CI:0, capturing fingerprints during the early morning, for the offline phase of our experiments. This means that the training process solely relied on this subset of data from CI:0. The remaining data from CI:0 and CIs:1-9 were reserved for testing purposes. In the dataset, we split the data in a 5:1 ratio to ensure a robust evaluation of the model's performance. During the training phase, we utilized 5 fingerprints per RP from the OP3 device to train the GM, and we used 1 fingerprint per RP from the same OP3 device to train the SSM. For experiments evaluating the effects of model poisoning attacks, we consistently designated the MOTO device as the adversarial client. This device was configured to generate malicious LM updates across all model poisoning related experiments to ensure consistency.

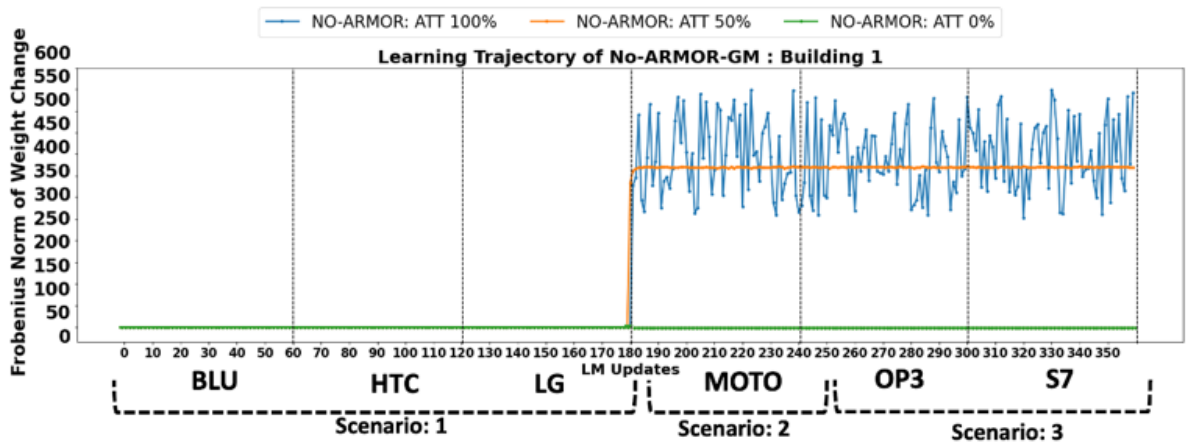

(a)

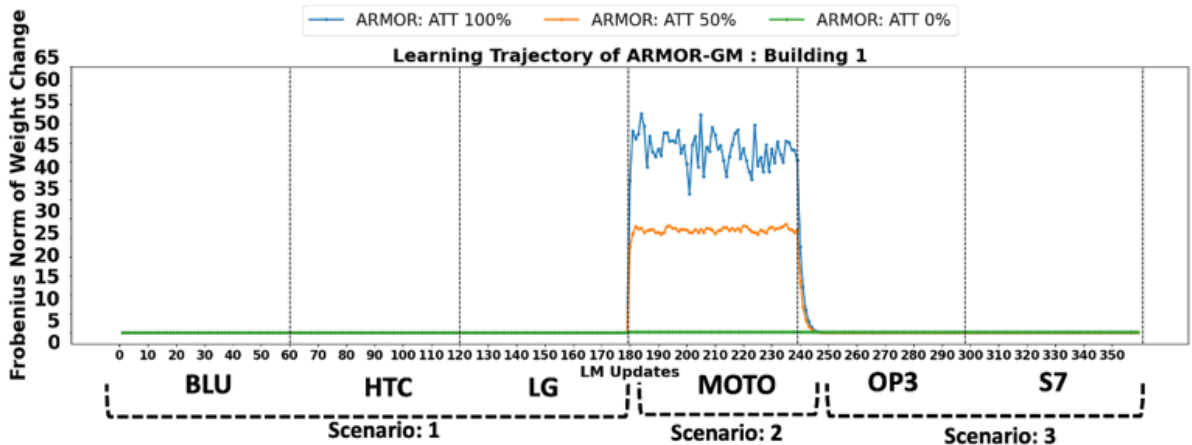

(b)

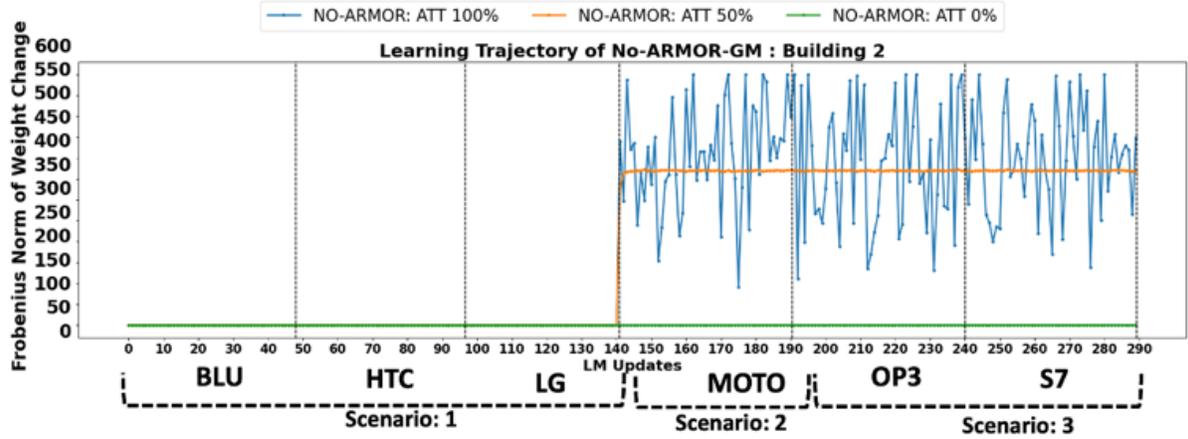

(c)

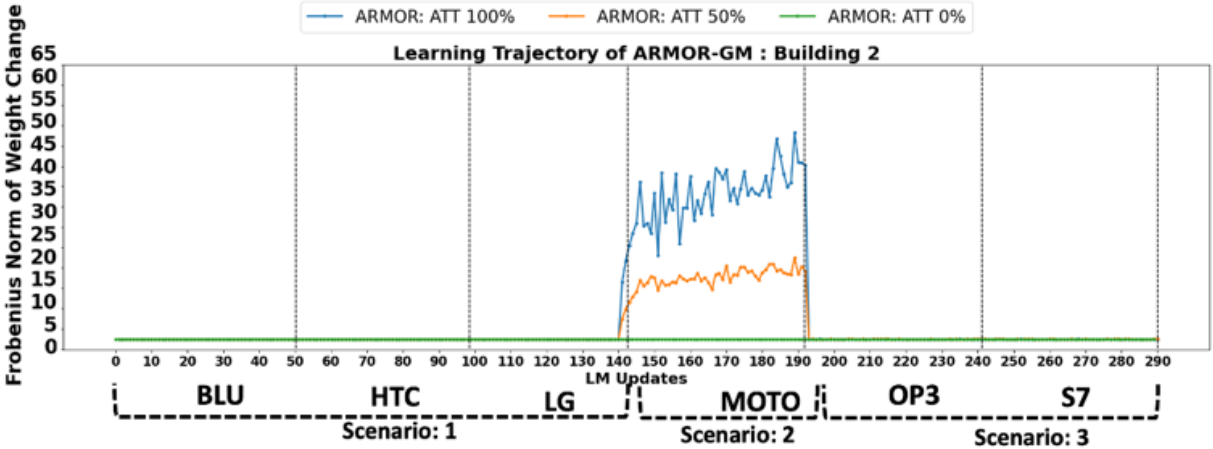

(d)

Figure 6. Learning trajectory of GM weights in (a) Building 1 for No-ARMOR-GM, (b) Building 1 for ARMOR-GM, (c) Building 2 for No-ARMOR-GM, (d) Building 2 for ARMOR-GM.

### 6.3 Effects of Model Corruption on Learning Trajectory

We first investigate the changes in the learning trajectory of the GM with and without the presence of model corruption, utilizing the *CSUIndoorLoc* dataset with CI:0, as illustrated in Figures 6(a) through 6(d). In the figures, the X-axis denotes the number of LM updates submitted by each of the client devices, while the Y-axis denotes the Frobenius norm of weight change. This Frobenius norm indicates the changes in the GM weight tensors concerning the previous GM state, providing insight into the GM's learning process within the CFL setting. A low Frobenius norm value implies minor weight changes and a stable learning trajectory, while a high value suggests significant weight changes, indicating potential instability or corruption in the learning process.

We compare the learning trajectories of the GM in the ARMOR framework (ARMOR-GM) against a baseline GM without ARMOR's specialized corruption mitigation strategies (No-ARMOR-GM). The No-ARMOR-GM uses the same GM parameters as ARMOR-GM but lacks the corruption mitigation mechanisms integral to ARMOR. In each of the figures (6 (a) through 6 (d)), we analyze three distinct scenarios to evaluate the impact of model corruption.

In the first scenario, we evaluate the learning trajectory without any model poisoning attacks. The LM updates submitted by the BLU, HTC, and LG devices are received by both the ARMOR-GM and No-ARMOR-GM without any model poisoning interference. For both Building 1 (as illustrated in Figure 6 (a) and 6 (b)) and Building 2 (as illustrated in Figure 6 (c) and 6 (d)), the Frobenius norm for ARMOR-GM and No-ARMOR-GM remains low and flat, indicating minimal weight changes and a stable learning trajectory. The low Frobenius norm values imply that the model weights are not experiencing substantial changes, which is characteristic of a healthy learning process with consistent updates.

In the second scenario, we introduce model poisoning attacks using the MOTO device configured to submit the Gaussian attack at different attack strengths (ATT) of 0%, 50%, and 100%. At *ATT* 0%, the learning trajectory for both ARMOR-GM and No-ARMOR-GM remains largely unchanged from the first scenario, maintaining a low and flat Frobenius norm, indicating that there is no impact on learning from benign updates. However, at *ATT* 50% and 100%, a significant divergence between ARMOR-GM and No-ARMOR-GM becomes evident. The No-ARMOR-GM shows a sharp increase in the Frobenius norm, suggesting that the model's weight updates have become more erratic due to the introduction of poisoned updates. The increase in Frobenius norm indicates significant changes in the model weights, reflecting its susceptibility to corruption. In contrast, ARMOR-GM shows a more moderated increase in the Frobenius norm under *ATT* 50% and 100%. While there is an uptick due to the introduction of poisoned updates, the rise is more controlled, and the values stabilize more quickly. This indicates that ARMOR's corruption mitigation strategies effectively dampen the impact of the poisoned updates, reducing their effect on the model's learning process. The reason ARMOR-GM displays a smaller increase in the Frobenius norm compared to No-ARMOR-GM lies in its adaptive federated aggregation strategy, specifically designed to mitigate corruption.

In the third scenario, after the model poisoning updates, we reintroduce legitimate client updates to observe the recovery behavior of both models. The No-ARMOR-GM struggles to recover from the impact of the poisoned updates. The Frobenius norm, remains relatively high compared to its initial values before the attack. This suggests that the model's weights continue to reflect the impact of earlier poisoning attacks, leading to potential long-term degradation in model performance. The lack of effective mitigation strategies in No-ARMOR-GM prevents it from fully purging the corrupted information from its system, resulting in a lingering effect that hampers its learning stability. In contrast, ARMOR-GM demonstrates a marked recovery following the reintroduction of legitimate updates. The Frobenius norm gradually decreases towards its pre-attack levels, indicating that the model is effectively filtering out the corrupted information and stabilizing back to a normal learning trajectory.

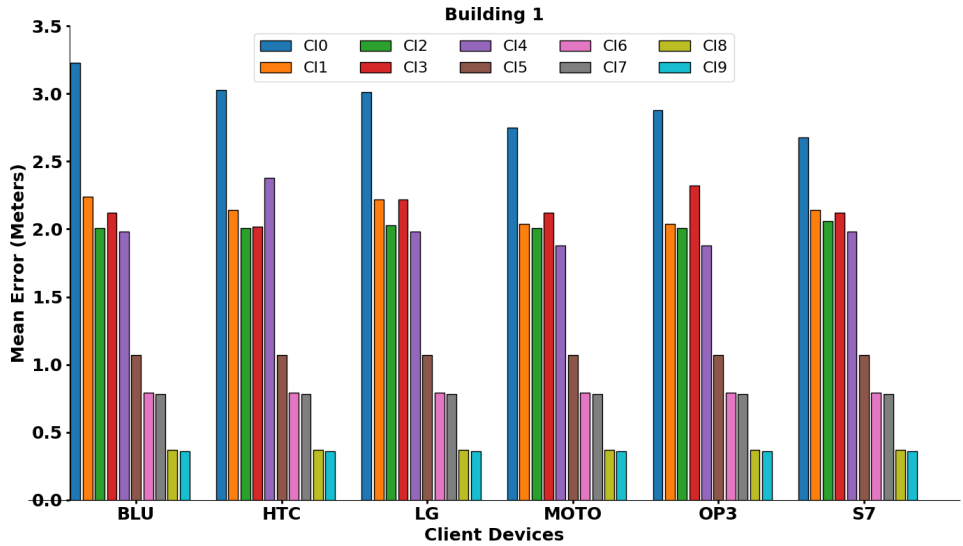

(a)

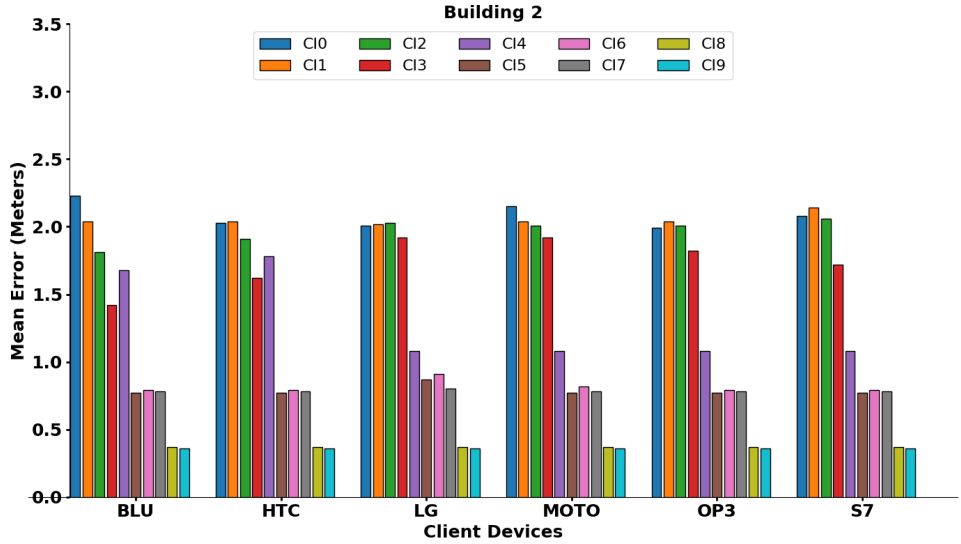

(b)

Figure 7. Mean localization error for different client devices and CIs in (a) Building 1 and (b) Building 2.

*6.4 Effects of Device Heterogeneity and Temporal Fluctuations*

Next, we present the performance of the ARMOR framework under RSS fluctuations induced by both device heterogeneity and temporal environment dynamics. Bar plots for Building 1 and Building 2 are shown in Figure 7(a) and Figure 7 (b) respectively, where the x-axis denotes the client devices, and the y-axis denotes the mean localization error in meters. Each bar represents the mean localization error across all RPs for a particular client device without introducing any model poisoning attacks. Different colored bars denote the mean error for each CI, ranging from CI:0 to CI:9.

In Building 1 as shown in Figure 7 (a), the mean localization errors across different devices are shown. It is evident that the mean errors decrease over successive CIs, indicating that the ARMOR framework can incrementally learn and adapt to the temporal dynamics of the environment. The initial CIs (CI:0) show higher mean errors, particularly for the BLU and HTC devices, suggesting that the framework initially struggles to adapt to the inherent heterogeneity and noise present in the early data. However, as more data is collected and processed (up to CI:9), the mean errors reduce significantly, demonstrating the framework's capability to improve localization accuracy over time.

Similarly, in Building 2 as shown in Figure 7 (b), the trend of decreasing mean localization errors with successive CIs is observed across all client devices. The BLU and LG devices show relatively higher initial mean errors, which decrease as the ARMOR framework processes more data. The analysis reveals that in Building 1, the mean localization errors reduce by approximately 90.3% to 93.3% from CI:0 to CI:9. In Building 2, the mean errors reduce by approximately 88.9% to 92.3% over the same period. This significant reduction in errors highlights the ARMOR framework's ability to incrementally learn and improve accuracy despite the challenges posed by varying device characteristics and environmental conditions over time.

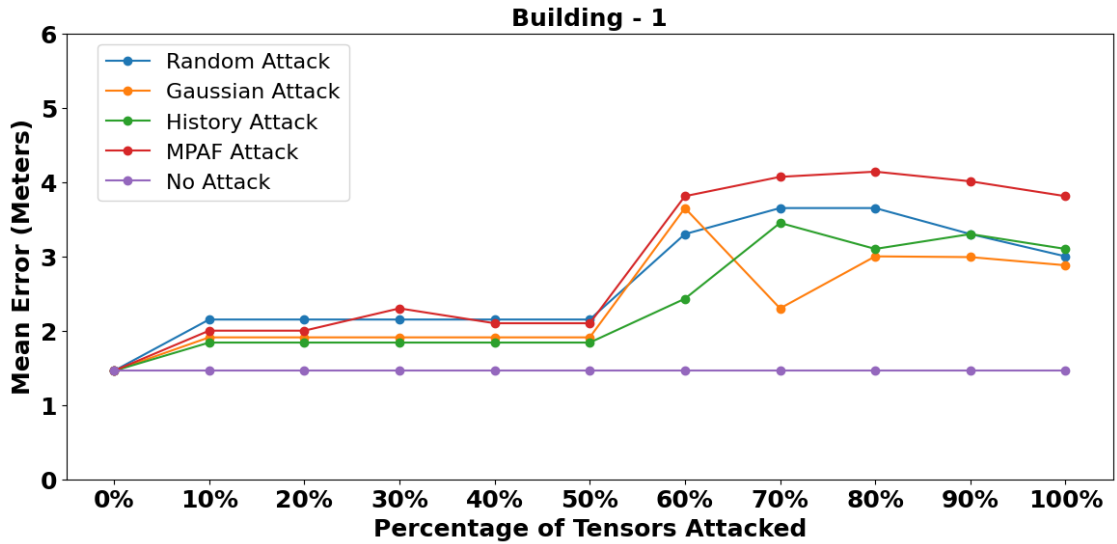

(a)

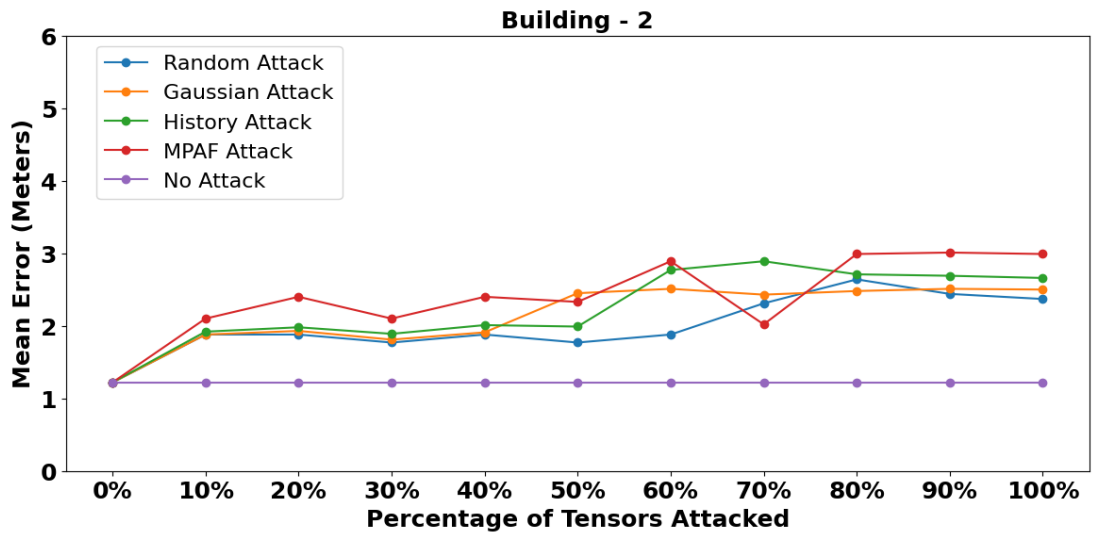

(b)

Figure 8. Mean localization error for different attack strengths (*ATT*) and model poisoning attacks in (a) Building 1 and (b) Building 2.

## 6.5 Effects of Model Poisoning Attacks in CL

Next, we analyze the impact of various model poisoning attacks on the mean localization error in the CFL setting, using the ARMOR framework. Figures 8 (a) and 8 (b) shows the mean localization error across different attack strengths (*ATT*), with four different types of attacks: Random Attack, Gaussian Attack, History Attack, and MPAF Attack. The x-axis denotes the *ATT*, while the y-axis shows the mean localization error in meters. We present separate plots for Building 1 (Figure 8 (a)) and Building 2 (Figure 8 (b)).

To maintain consistency with the analysis from Section 6.3, we selected the MOTO device to submit model corruptions. The results in Building 1 demonstrate the ARMOR framework's resilience against all model poisoning attacks, even under potent conditions with high *ATT*. Although the mean localization error generally increases as the *ATT* increases, the ARMOR framework manages to maintain reasonable localization accuracy. The MPAF Attack results in the highest mean error, peaking at around 80% *ATT*, indicating its potency. However, even with such high attack intensity, the increase in mean error is controlled, showcasing the robustness of the ARMOR framework. The Gaussian Attack and Random Attack follow a similar trend, with mean errors increasing slightly beyond 50% *ATT*. The History Attack, however, shows more resilience, maintaining a relatively lower mean error compared to the other attacks, even at higher *ATT*.

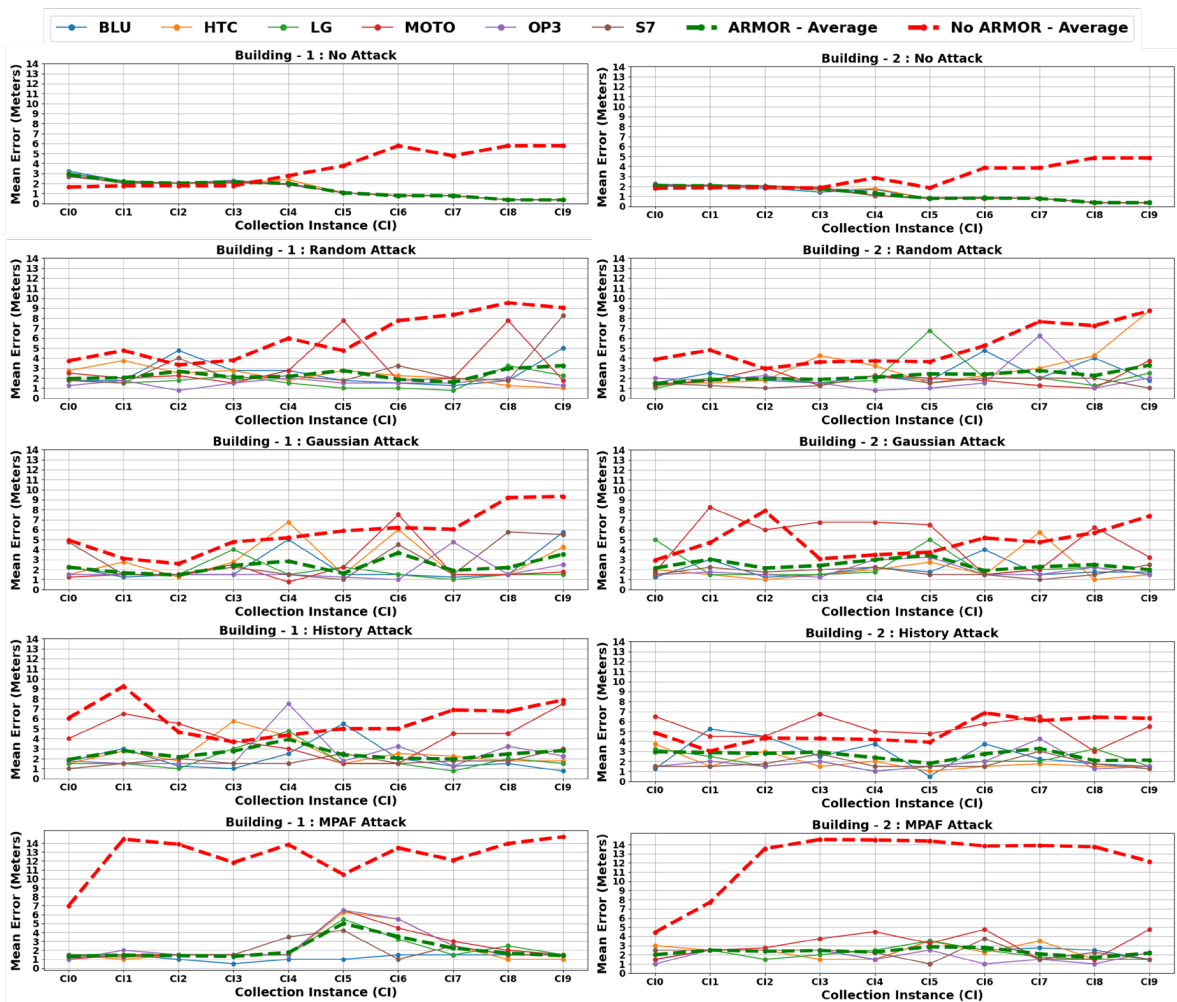

Figure 9. Mean localization error under various model poisoning attacks across different CIs for Building 1 and Building 2.

In Building 2, the trend is consistent with Building 1, however, the overall mean errors are slightly lower compared to Building 1, suggesting that the ARMOR framework performs marginally better in this environment, potentially due to the smaller path length of Building 2. The MPAF Attack again leads to the highest errors, especially noticeable at 80% *ATT*. The Gaussian and Random Attacks show a more consistent increase in error, while the History Attack maintains the lowest mean error, showcasing its effectiveness in mitigating the impact of model poisoning. It is important to note that in real-world conditions, adversaries typically employ lower *ATTs* to stay covert. This makes the ARMOR framework's resilience even more significant, as it demonstrates the capability to mitigate the effects of model corruption induced by deliberate attacks even under extreme conditions.

Furthermore, to evaluate the performance of ARMOR under different model poisoning attacks across multiple time periods—represented as collection instances (CIs), we present the results in Figure 9. The x-axis represents the sequential CIs – where CI0–CI2 were collected on the same day with 6-hour intervals. CI3 was collected after 24 hours, CI4 after 7 days, CI5 after 30 days, CI6 after 2 months, CI7 after 3 months, CI8 after 4 months, and CI9 after 8 months. The y-axis denotes the mean localization error (in meters) averaged across all RPs for each building floorplan. We first examine the 'No Attack' condition to isolate the effect of temporal dynamics and without any model poisoning attacks in the indoor environment. For each CI, we show the localization error per device as well as the average error across all devices. For comparison, we include a 'No ARMOR – Average' curve, which represents the average localization error of all devices per CI when ARMOR's model corruption monitoring, detection, and mitigation mechanisms are disabled. This configuration effectively reduces the system to a conventional CFL setup without corruption safeguarding. Under the no-attack setting, we observe that the No ARMOR variant exhibits gradual performance degradation as the CIs progress, indicating susceptibility to temporal dynamics and accumulated erroneous LM updates. Notable spikes are observed after CI4 for Building-1 and CI5 for Building-2. In contrast, ARMOR (shown as 'ARMOR-Average' in Figure 9) maintains lower errors across all CIs. This behavior demonstrates that the SSM used within ARMOR successfully tracks the GM's learning trajectory and mitigates corruptions caused by temporally inconsistent or biased LM updates.

We then evaluate ARMOR under four model poisoning attacks: Random, Gaussian, History, and MPAF attacks. To maintain consistency with the analysis in Figures 8(a) and 8(b), the MOTO device is selected as the model poisoning participant within each CI, submitting poisoned updates across all RPs, while all other devices continue to provide benign updates and while the set *ATT* to 100% (indicating all weight tensor being manipulated by the attack). This setup reflects a scenario in which a single compromised device can influence the CFL aggregation process over time. Across all attack types and both buildings, the No-ARMOR variant consistently exhibits significantly higher localization errors and a pronounced upward error trend across CIs. This demonstrates that model poisoning attacks, when combined with temporal dynamics, can severely degrade a conventional CFL-based indoor localization system—even when only a single device is poisoned. In several cases, the No-ARMOR variant reaches error levels approaching 15 meters.

In contrast, ARMOR consistently maintains lower localization errors across all CIs and attack scenarios. Even under potent attacks and extended temporal variations (spanning up to 8 months of data collection – CI9), ARMOR maintains the average localization error below 6 meters for Building-1 and below 4 meters for Building-2. This stability confirms that the SSM-driven corruption monitoring, detection, and mitigation phases effectively safeguards the GM from corrupted LM updates and preserves localization accuracy over time.

*6.6 Comparison Against State-of-the-Art*

Lastly, we compare the ARMOR framework's performance against several state-of-the-art methods, including KRUM [45], Multi-KRUM [45], Bulyan [46], FedLoc [41], and FedHIL [42]. KRUM [45] selects the LM update closest to the majority in the GM history for aggregation, Multi-KRUM [45] extends KRUM by aggregating multiple such LM updates, and Bulyan [46] performs LM outlier filtering before GM aggregation. These methods are designed to defend against model poisoning attacks in FL and are included to evaluate ARMOR's robustness under adversarial conditions. FedLoc [41] employs FedSGD for privacy-preserving localization, while FedHIL [42] addresses device heterogeneity through selective weight aggregation; both are included to benchmark ARMOR as domain-specific FL frameworks for real-world indoor localization.

The comparison is based on the mean localization error across all client devices, including the device inducing model poisoning attacks, averaged over all four attack methods and *ATTs* ranging from 0% to 100%. The results are depicted in Figure 10 (a) for Building 1 and Figure 10 (b) for Building 2, with the X-axis representing the different frameworks and the Y-axis showing the mean localization error in meters. In Figures 10 (a) and (b), the lower whisker represents the best-case scenario, showing the minimum localization error, while the upper whisker indicates the worst-case scenario, reflecting the maximum error observed, and the orange line within the box represents the average error.

For Building 1 as shown in Figure 10 (a), the ARMOR framework consistently achieves the lowest mean localization error with minimal variance, indicating its robustness and resilience against model corruption. The mean localization error for ARMOR remains well below 4 meters, while the other methods exhibit mean errors ranging from approximately 13.9 to 18.5 meters. Specifically, ARMOR demonstrates an average improvement of 5.71×, 6.86×, 6.75×, 7.25×, and 7.58× for Bulyan, FedHIL, Multi-KRUM, KRUM, and FedLoc respectively. In the worst-case scenario, ARMOR shows an improvement of approximately 3.36×, 3.06×, 3.13×, 3.71×, and 4.5× for Bulyan, FedHIL, Multi-KRUM, KRUM, and FedLoc respectively.

Similarly, in Building 2 as shown in Figure 10 (b), the ARMOR framework shows superior performance with a mean localization error significantly lower than the other methods. The ARMOR framework's mean error remains below 3 meters, while the other methods show mean errors ranging from approximately 10.44 to 15.52 meters. The average improvement in mean localization error for ARMOR is 5.38×, 5.77×, 6.06×, 6.7×, and 8.0× for Bulyan, FedHIL, Multi-KRUM, KRUM, and FedLoc respectively. In the worst-case scenario, ARMOR shows an improvement of approximately 3.72×, 2.86×, 3.75 ×, 4.97×, and 4.69× for Bulyan, FedHIL, Multi-KRUM, KRUM, and FedLoc respectively.

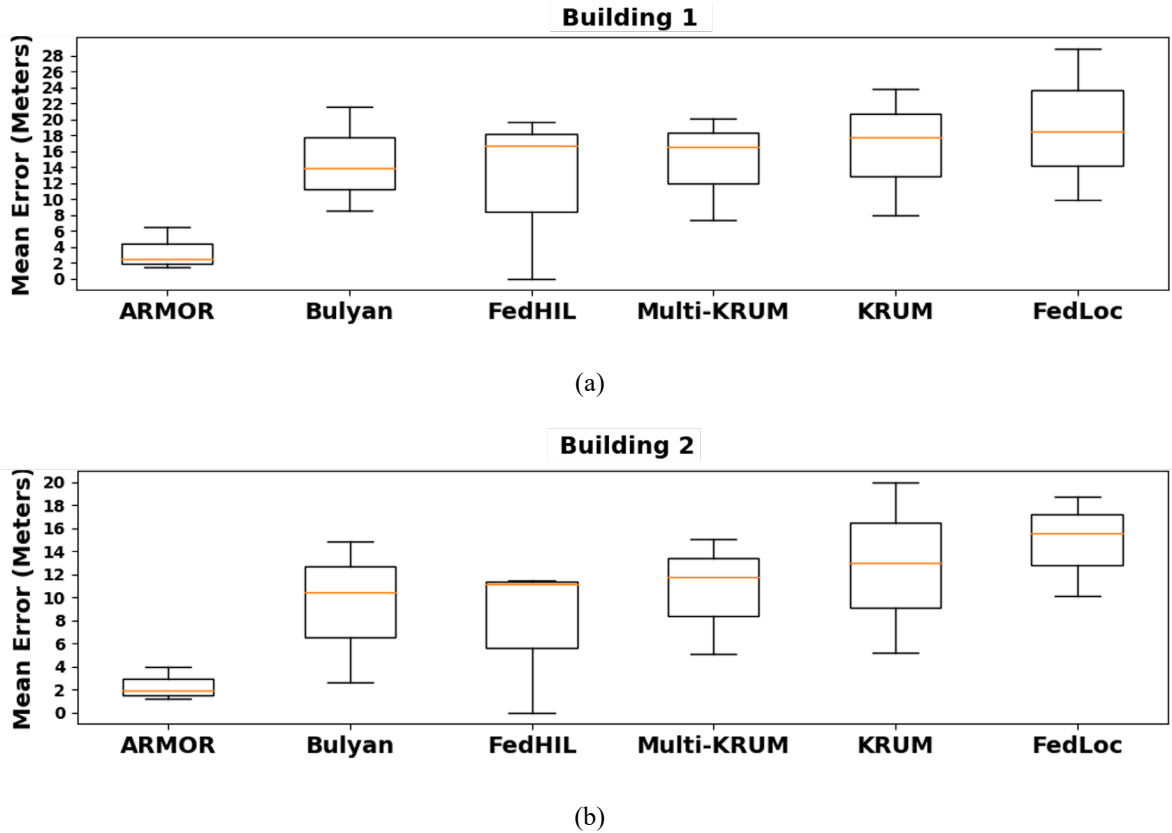

(a)

(b)

Figure 10. Comparison of mean localization errors for ARMOR and state-of-the-art methods across all attack methods and *ATT* in (a) Building 1 and (b) Building 2.

The advantage of ARMOR over other frameworks lies in its ability to manage corrupted updates more effectively. ARMOR's phases of corruption monitoring, detection, and mitigation ensure that significant changes to the GM from malicious client updates are minimized, which leads to lower localization errors. In contrast, Bulyan, though effective at filtering extreme outliers, often discards legitimate updates along with them, negatively affecting its overall performance. FedHIL's selective weight aggregation performs well in best-case scenarios by prioritizing relevant LM tensors, but it also inadvertently aggregates a subset of corrupted tensors, making it less effective in more sophisticated poisoning attacks, resulting in higher worst-case and average errors. Multi-KRUM and KRUM, while designed to defend against model poisoning attacks by selecting central updates, are less capable in situations where subtle manipulations remain undetected, leading to higher localization errors. Finally, FedLoc, although effective in non-poisoning conditions, lacks robust anomaly detection, leaving it vulnerable to model poisoning, resulting in the highest mean and worst-case errors among the methods.

## 7. CONCLUSION

In this article, we introduced the ARMOR framework, designed to enhance the resilience and accuracy of CFL indoor localization systems against model corruption induced by RSS fluctuations and model poisoning attacks. Our comprehensive evaluation using the open-source *CSUIndoorLoc* dataset demonstrated the ARMOR framework's robust performance in real-

23world scenarios, highlighting its ability to maintain a stable learning trajectory and accurate localization despite various challenges. Our experiments revealed that the ARMOR framework effectively mitigates the impact of model corruption, achieving much lower mean localization errors compared to traditional FL-based frameworks and other state-of-the-art methods. Additionally, the framework showed remarkable resilience, with an average improvement of up to 8.0× in mean errors and up to 4.97× in worst-case errors across both buildings compared to the best-known prior works.

While this study focused on model corruption driven by temporal dynamics and model poisoning attacks, data poisoning attacks represent another important threat in CFL systems. Future work could investigate how ARMOR's learning trajectory monitoring method can be extended to settings where malicious behavior originates at the data level, enabling a more comprehensive defense strategy for CFL-based indoor localization.

**References**

[1] *MarketsandMarkets*. "Indoor Location Market by Component, Technology, Application, Vertical and Region - Global Forecast to 2028." 2024 [Online], https://www.marketsandmarkets.com/Market-Reports/indoor-location-market-989.html.
[2] Google. "Indoor Maps." 2011 [Online], https://www.google.com/maps/about/partners/indoormaps
[3] Apple. "Apple Indoor Maps and Positioning." 2021 [Online], https://register.apple.com/resources/indoor/program
[4] Z. Zeng, L. Wang, and S. Liu. "An introduction for the indoor localization systems and the position estimation algorithms." *IEEE WorldS4*, 2019.
[5] PM. Rebelo, J. Lima, SP. Soares, PM. Oliveira, H. Sobreira, and P. Costa. "A Performance Comparison between Different Industrial Real-Time Indoor Localization Systems for Mobile Platforms." *MDPI Sensors*, 2024.
[6] KW. Zegeye, SB. Amsalu, Y. Astatke, and F. Moazzami. "WiFi RSS fingerprinting indoor localization for mobile devices." *IEEE UEMCON*, 2016.
[7] S. Sadowski, and P. Spachos. "Rssi-based indoor localization with the internet of things." *IEEE Access*, 2018.
[8] C. Langlois, S. Tiku, and S. Pasricha. "Indoor localization with smartphones: Harnessing the sensor suite in your pocket." *IEEE CEM*, 2017.
[9] L. Li, X. Guo, and N. Ansari. "SmartLoc: Smart wireless indoor localization empowered by machine learning." *IEEE TIE*, 2019.
[10] N. Singh, S. Choe, and R. Punmiya. "Machine learning based indoor localization using Wi-Fi RSSI fingerprints: An overview." *IEEE Access*, 2021.
[11] S. Tiku, S. Pasricha, B. Notaros, and Q. Han. "A Hidden Markov Model based smartphone heterogeneity resilient portable indoor localization framework." *Journal of Systems Architecture*, 2020.
[12] M. Zhou, Y. Li, MJ. Tahir, X. Geng, Y. Wang, and W. He. "Integrated statistical test of signal distributions and access point contributions for Wi-Fi indoor localization." *IEEE TVT*, 2021.
[13] DC. Nguyen, M. Ding, PN. Pathirana, A. Seneviratne, J. Li, and HV. Poor. "Federated learning for internet of things: A comprehensive survey." I*EEE CST*, 2021.
[14] H. Zhu, J. Xu, S. Liu, and Y. Jin. "Federated learning on non-IID data: A survey." *Neurocomputing*, 2021.
[15] Y. Etiabi, and EM. Amhoud. "Federated distillation based indoor localization for IoT networks." *IEEE Sensors*, 2024.
[16] Y. Li, J. Ibrahim, H. Chen, D. Yuan, and KKR. Choo. "Holistic Evaluation Metrics: Use Case Sensitive Evaluation Metrics for Federated Learning." *arXiv preprint arXiv:2405.02360*, 2024.
[17] O. Tasbaz, B. Farahani, and V. Moghtadaiee. "Feature fusion federated learning for privacy-aware indoor localization." *Peer-to-Peer Networking and Applications,* 2024.
[18] G. Graffieti, G. Borghi, and D. Maltoni. "Continual learning in real-life applications." *IEEE RAL*, 2022.
[19] K. Shaheen, MA. Hanif, O. Hasan, and M. Shafique. "Continual learning for real-world autonomous systems: Algorithms, challenges and frameworks." *Journal of Intelligent & Robotic Systems*, 2022.
[20] L. Wang, X. Zhang, H. Su, and J. Zhu. "A comprehensive survey of continual learning: theory, method and application." *IEEE TPAML*, 2024.
[21] I. Ashraf, S. Hur, and Y. Park. "Smartphone sensor based indoor positioning: Current status, opportunities, and future challenges." *MDPI Electronics*, 2020.
[22] K. Pillutla, SM. Kakade, and Z. Harchaoui. "Robust aggregation for federated learning." *IEEE TSP*, 2022.
[23] N. Bouacida, and P. Mohapatra. "Vulnerabilities in federated learning." *IEEE Access*, 2021.
[24] Y. Zuo, L. Gui, K. Cui, J. Guo, F. Xiao, and S. Jin. "Mobile Blockchain-Enabled Secure and Efficient Information Management for Indoor Positioning With Federated Learning." *IEEE TMC*, 2024.
[25] F. Potorti, S. Park, A. Crivello, F. Palumbo, M. Girolami, P. Barsocchi, S. Lee, J. Torres-Sospedra, ARJ. Ruiz, A. Perez-Navarro, and GM. Mendoza-Silva. "The IPIN 2019 indoor localisation competition—Description and results." *IEEE Access*, 2020.
[26] D. Lymberopoulos and J. Liu. "The microsoft indoor localization competition: Experiences." *IEEE SPM*, 2017
[27] G. Chen, X. Guo, K. Liu, X. Li, and J. Yang. "RWKNN: A modified WKNN algorithm specific for the indoor localization problem." *IEEE Sensors Journal,* 2022.
[28] MT. Le. "Enhanced indoor localization based BLE using Gaussian process regression and improved weighted kNN." *IEEE Access*, 2021.